\documentclass{article}

\usepackage{arxiv}

\usepackage{hyperref}
\usepackage{algorithm}
\usepackage{algpseudocode}

\usepackage{amssymb}
\usepackage{float}

\usepackage[T1]{fontenc}
\usepackage{graphicx}
\usepackage{subfig}
\usepackage{booktabs}
\usepackage[misc]{ifsym}   

\usepackage{amsmath}
\usepackage{amssymb}

\usepackage{xcolor}
\hypersetup{
  colorlinks=true,
  linkcolor=blue,
  citecolor=blue,
  urlcolor=blue
}

\usepackage[utf8]{inputenc} 
\usepackage[T1]{fontenc}    
\usepackage{hyperref}       
\usepackage{url}            
\usepackage{booktabs}       
\usepackage{amsfonts}       
\usepackage{nicefrac}       
\usepackage{microtype}      
\usepackage{lipsum}
\usepackage{graphicx}
\graphicspath{ {./images/} }

\title{DOVA-PATBM: An Intelligent, Adaptive, and Scalable Framework for Optimizing Large-Scale EV Charging Infrastructure}

\author{
 Chuan Li \\
 LIPADE, Université Paris Cité \\
 Sorbonne Université \\
 SAMOVAR, Telecom SudParis\\
 Institut Polytechnique de Paris \\
 Paris, France \\
 \texttt{chuan.li@sorbonne-universite.fr} \\
 \And
 Shunyu Zhao \\
 The Institute of Statistical Mathematics \\
 The Graduate University for Advanced Studies \\
 Tokyo, Japan \\
 \texttt{zhao.shunyu@ism.ac.jp} \\
 \And
 Vincent Gauthier \\
 SAMOVAR, Telecom SudParis \\
 Institut Polytechnique de Paris \\
 Palaiseau, France \\
 \texttt{vincent.gauthier@telecom-sudparis.eu} \\
 \And
 Hassine Moungla \\
 LIPADE, Université Paris Cité \\
 SAMOVAR, Telecom SudParis\\
 Institut Polytechnique de Paris \\
 Paris, France \\
 \texttt{hassine.moungla@u-paris.fr} \\
}

\begin{document}
\maketitle
\begin{abstract}
The accelerating uptake of battery–electric vehicles demands
infrastructure planning tools that are both \emph{data–rich} and
\emph{geographically scalable}.  Whereas most prior studies optimise
charging locations for single cities, state-wide and national networks
must reconcile the conflicting requirements of dense metropolitan cores,
car-dependent exurbs, and power-constrained rural corridors.  We present
\textbf{DOVA-PATBM}—\textit{Deployment Optimisation with
Voronoi-oriented, Adaptive, POI-Aware Temporal Behaviour Model}—a
geo-computational framework that unifies these contexts in a single
pipeline.  The method rasterises heterogeneous data (roads, population,
night lights, POIs, and feeder lines) onto a hierarchical H3 grid,
infers intersection importance with a zone-normalised
graph-neural-network centrality model, and overlays a Voronoi tessella­
tion that guarantees at least one five-port DC fast charger within every
30 km radius.  Hourly arrival profiles, learned from loop-detector and
floating-car traces, feed a finite M/M/\(c\) queue to size ports under
feeder-capacity and outage-risk constraints.  A greedy
maximal-coverage heuristic with income-weighted penalties then selects
the minimum number of sites that satisfy coverage and equity targets.

Applied to the State of Georgia, USA, DOVA-PATBM (i) increases 30 km
tile coverage by 12 percentage points, (ii) halves the mean distance
that low-income residents travel to the nearest charger, and (iii)
meets sub-transmission headroom everywhere—all while remaining
computationally tractable for national-scale roll-outs.  These results
demonstrate that a tightly integrated, GNN-driven, multi-resolution
approach can bridge the gap between academic optimisation and deployable
infrastructure policy.
\end{abstract}


\maketitle
\begin{figure}[htbp]
  \centering
  \includegraphics[width=\linewidth]{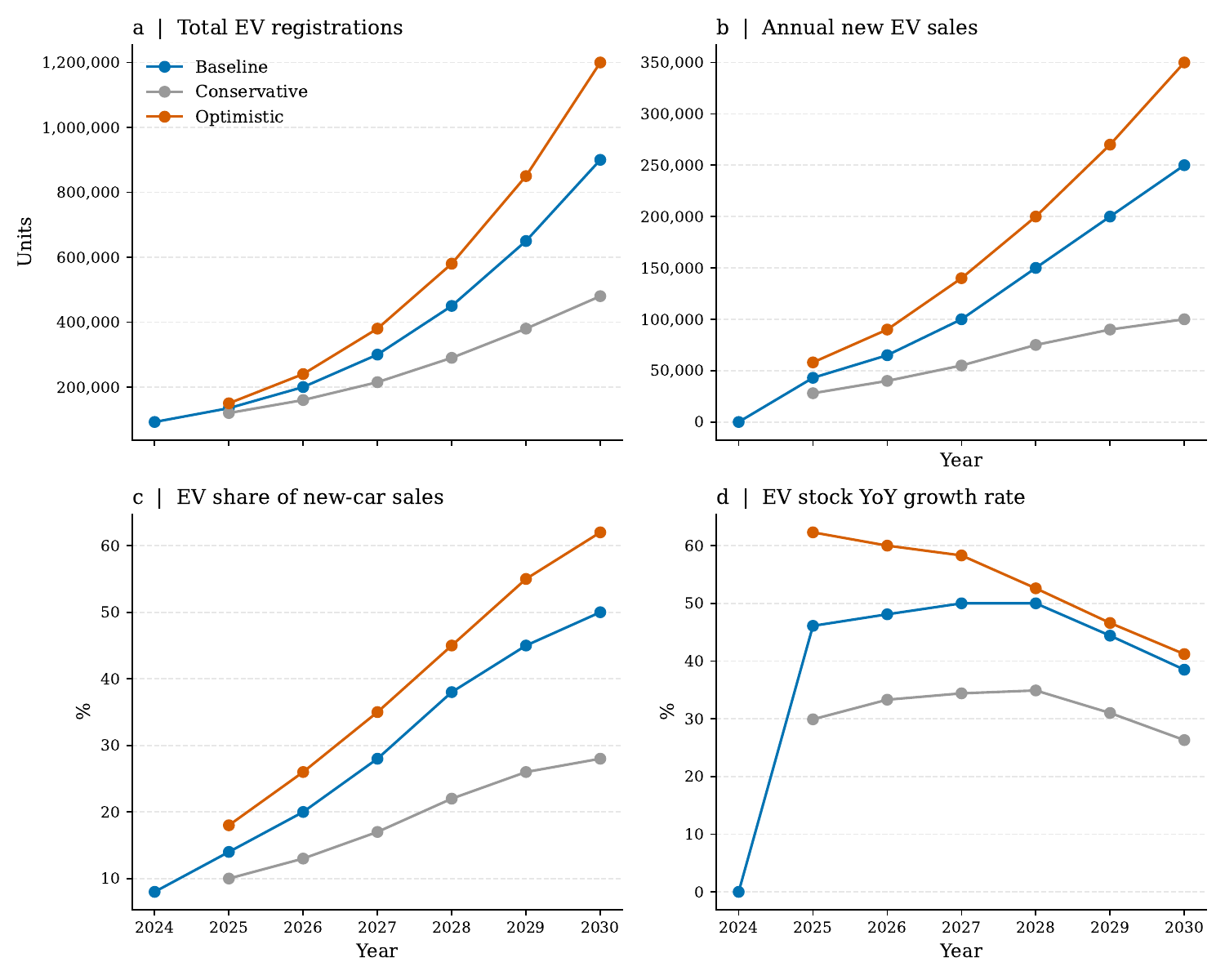}
  \caption{Geogia EV tendance}
  \label{fig:ga_state_ev}
\end{figure}

\section{Introduction}

\begin{figure*}[ht]
  \centering
  \includegraphics[width=\textwidth]{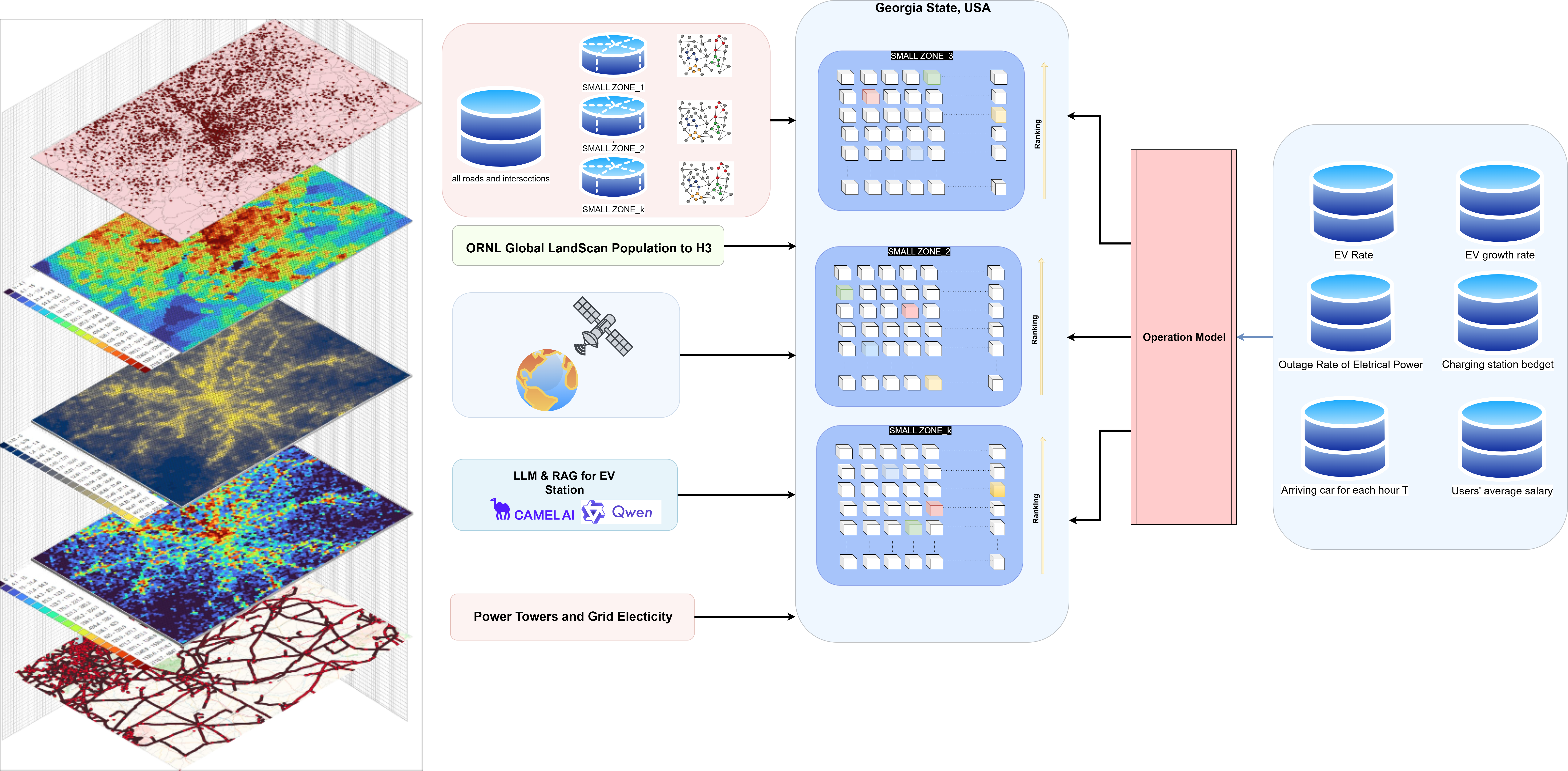}
  \caption{Schematic of DOVA-PATBM.  
  \textbf{Left:} multi-layer spatial inputs—road network and
  intersections, LandScan population raster, VIIRS night-light
  intensity, POI densities extracted with a Qwen / CAMEL-AI
  zero-shot pipeline, and transmission-line topology.  
  \textbf{Centre:} Georgia is partitioned into a nested
  \texttt{H3\_6}\,$\rightarrow$\,\texttt{H3\_8} grid.  Within each cell
  a zone-normalised GNN assigns intersection centrality; cells are then
  Voronoi-segmented to enforce a 30 km reachability guarantee.  
  \textbf{Right:} the operations module sizes chargers using a
  finite-capacity M/M/$c$ queue that ingests hourly arrival rates,
  budget limits, feeder headroom, outage risk, and income-weighted
  fairness targets.  The optimiser outputs a geo-referenced build plan with port counts and phased budgets.}
  \label{fig:dova_patbm_workflow}
\end{figure*}

The accelerating electrification of Georgia’s vehicle fleet provides an
urgent backdrop for infrastructure planning.  Official AFDC statistics
reveal that battery-electric stock in the state more than doubled—from
\(\approx 47{\,}000\) units in 2021 to 92{\,}000 at the end of 2023
\cite{DOE_AFDC_Registrations2024}.  The Atlanta Regional Commission
expects electric vehicles to capture  30–60 \%  of new-car sales in
Metro Atlanta by 2030 \cite{ARC_RTEP2025}, a trajectory that translates,
even under conservative assumptions, into  0.48 million  registered
EVs statewide by the end of the decade; an optimistic scenario exceeds
 1.2 million  \cite{AFDC_MapsData2025,PublicNewsService2023}.  Such
growth implies an order-of-magnitude increase in public charging demand,
particularly in urban cores where adoption starts highest
\cite{morocho2024identifying} and in rural counties that currently have
minimal provision \cite{10136865}.  Meeting that demand in an equitable,
power-constrained, and budget-aware manner motivates the present work.

We introduce \textbf{DOVA-PATBM}—\textit{Deployment Optimisation with
Voronoi-oriented, Adaptive, POI-Aware Temporal Behaviour Model}—a
fully-automated siting framework whose processing chain is summarised in
Fig.~\ref{fig:dova_patbm_workflow}.  Heterogeneous data layers
(transport, demographics, night lights, POIs, and grid assets) are first
rasterised to a multi-resolution H3 mesh.  A graph neural network
provides an inductive, zone-normalised centrality score for every road
intersection.  Urban cells (\texttt{H3\_8}) retain their native
2km grain, whereas suburban and rural territory is coarsened to
\texttt{H3\_7} and \texttt{H3\_6}; a Voronoi overlay then ensures that a
five-port DC fast charger lies within 30km of any point in the state.
Temporal modules learn hourly arrival profiles from loop detectors and
floating-car traces, enabling an outage-aware M/M/$c$ queue to right-size
each site.  Finally, a greedy maximal-coverage heuristic with
income-weighted penalties selects the minimal set of stations that
satisfies both coverage and equity targets.

Applied to Georgia, DOVA-PATBM delivers a \textbf{12pp} improvement in
tile coverage at a 30km service radius, cuts the average
low-income–resident travel distance to the nearest charger by
\textbf{50\%}, and respects sub-transmission headroom everywhere.  These
results underscore the framework’s ability to balance grid constraints,
behavioural heterogeneity, and social equity at state scale while
remaining computationally tractable for national roll-outs.

Building on our earlier proof-of-concept, which relied on fixed
population–traffic score weights to rank candidate cells
\cite{li2024large}, the present version of DOVA-PATBM adds four pivotal
enhancements.  First, the heuristic centrality metric is replaced by an
\emph{inductive} GraphSAGE model whose parameters are learned from a
stratified sample of county road graphs and then transferred to all
\(282\,771\) \texttt{H3\_8} cells.  The model ingests node degree,
latitude–longitude encodings, and POI densities, and outputs a
zone-normalised importance score that is robust to inter-zone size
variance.  Second, a hierarchical H3–Voronoi mesh dynamically coarsens
from 2km urban hexes to 14km rural cells; a Voronoi overlay guarantees
that a five-port DCFC hub lies within 30km of any point in the state.
Third, charger sizing is driven by a finite M/M/\(c\) queue that
incorporates \emph{(i)} an empirically observed outage rate,
\emph{(ii)} feeder headroom constraints, and \emph{(iii)} a
\textbf{salary-weighted waiting-cost term}.  The latter multiplies the
mean queue time by the county-level average hourly wage reported in the
Bureau of Labor Statistics, thereby steering capacity upgrades towards
high-wage employment centres where the economic cost of delay is
greatest.

Finally, DOVA-PATBM forecasts demand through 2030.  Annual EV stock and
new-sales shares are projected with a compound-growth model calibrated
to AFDC registrations and Atlanta Regional Commission targets; the
resulting county-year penetration surfaces update the arrival-rate
matrix \(\lambda_{i,t}^{(\text{year})}\) so that the optimiser can stage
deployments and capital outlays across a five-year horizon.  All
up-stream data—roads, population, night lights, POIs, grid assets—are
ingested via large-language-model pipelines (Qwen + CAMEL-AI) that
classify and score 380k+ POIs in under ten minutes on a single GPU,
ensuring the framework remains fully automated and reproducible.  The
combined improvements yield a deployable blueprint that respects power
constraints, captures behavioural heterogeneity, internalises labour
opportunity costs, and adapts to the rapid, county-specific growth in
EV adoption projected for the next decade.

\section{Related Work}

Early studies on EV-charger siting relied on multi-criteria
decision-making (MCDM) grids \cite{iravani2022multicriteria} and mixed-integer
programming formulations that optimise either the charging-station
owner’s profit, the distribution-network operator’s reliability, or the
user’s travel effort
\cite{chen2021optimal,ren2019location,othman2020optimal,yu2022map}.
While mathematically rigorous, these approaches scale poorly beyond
moderate-sized urban areas.  Two recent surveys provide exhaustive
overviews \cite{shen2019optimization,kchaou2021charging}.

To alleviate scalability constraints, researchers have turned to machine
learning and network science.  Graph-convolutional networks predict
spatio-temporal charging demand \cite{Wang2023predicting}, and
zero-shot LLMs classify POIs for candidate generation
\cite{Qu2024chatev,Feng2024large}.  Reinforcement-learning agents have
been trained to recommend fast-charger locations in dynamic traffic
networks \cite{Xu2022real}, whereas hierarchical cluster-and-rank
strategies suit low-density suburban contexts but struggle in congested
cores \cite{Lin2024hierarchical,Popa2024optimizing}.  At inter-city
scale, range-extension models map optimal stops on corridors
\cite{yilmaz2022range,li2024large}.

Queueing theory remains a critical component for sizing.  Classical work
combines M/M/\(c\) models with location problems
\cite{hosseini2015selecting,zhu2017planning}; more recent studies
propose bi-objective designs that incorporate grid impact and waiting
time yet still assume steady-state arrivals
\cite{meng2024research,wu2024allocate}.  Data-driven congestion
analytics at existing sites refine those assumptions
\cite{chen2017plug,xiao2020optimization}.

DOVA-PATBM advances the field on three fronts: (i) a GNN-based,
zone-normalised centrality metric that scales to 282 k \texttt{H3\_8}
cells; (ii) a nested H3–Voronoi mesh that unifies urban, suburban, and
rural siting without hand-tuned thresholds; and (iii) an integrated
queue-sizing module that respects feeder headroom and outage risk while
guaranteeing equity of access.  Together, these innovations close the
gap between academic optimisation and deployable, state-wide charging
strategy.

\section{Methodology}

To address the diversity in travel patterns, grid constraints, and data densities that characterize the \textit{urban}, \textit{suburban}, \textit{mixture suburban–rural}, and \textit{rural} regions of Georgia, we decompose the EV charger placement problem into two tightly-coupled stages. This hierarchical structure ensures adaptability to varying regional characteristics while preserving methodological consistency:

About fine-grained candidate generation,Each census block group is treated as a distinct \emph{zone}.  
We download the “drive” road network within each zone polygon using OSMnx, project it to an equal-area coordinate reference system (CRS), and construct a road graph. For every intersection node \(v\), we compute a \textit{degree-based GNN centrality score} \(c_v\), as defined in Algorithm~\ref{alg:centrality}.  
We retain only the intersections whose centrality values fall within the top \(\tau\)-th percentile (\(\tau{=}50\%\) by default) \emph{within the same zone}. This percentile-based filtering ensures that candidate sites are locally prominent within their respective zones, thus promoting geographic equity in siting.

For Location–allocation optimisation part.The retained candidate sites are connected to demand points—represented by population-weighted H3‐10 centroids enhanced with point-of-interest (POI) weighting—via a sparse distance-band coverage matrix.  
We then solve a \textit{maximal-coverage location problem (MCLP)} using either of two greedy heuristics:  
        (i) a \emph{budget-driven} heuristic that selects at most \(P\) sites to maximize coverage, or  
        (ii) a \emph{coverage-driven} heuristic that seeks the minimal number of sites to cover at least \(\alpha\!\times\!100\%\) of the total demand.  
        Additionally, when policy constraints require placing exactly one site per zone, we apply a \textit{per-zone scoring function} that linearly combines normalised centrality and estimated demand coverage using weights \(\beta_{\text{cent}}\) and \(\beta_{\text{cov}}\).

\subsection{GNN-based intersection centrality}
\label{subsec:gnn}

Let \(G_z = (V_z, E_z)\) denote the road graph extracted from zone \(z\), where nodes represent intersections and edges represent road segments. Each node is initialized with two types of features: (i) its degree \(d_v = |\mathcal{N}(v)|\), and (ii) positional encodings based on its geographic latitude and longitude.

We employ a two-layer GraphSAGE model with mean aggregation to learn centrality representations. The model is trained to predict \textit{betweenness centrality} targets computed from a small set of representative zones, and then deployed across all zones without fine-tuning. This approach balances computational efficiency with generalizability. The final learned centrality score for a node \(v\) is computed as:

\begin{equation}
  c_v = \sigma\!\left(\mathbf{W}_2\,
        \mathrm{ReLU}(\mathbf{W}_1\,h_v^{(1)})\right),
\end{equation}

where \(h_v^{(1)}\) is the hidden representation after one graph convolutional layer and \(\sigma\) denotes the sigmoid activation function. Since the model is applied independently to each zone, the resulting scores are only meaningful within the same zone—an intentional property that aligns with our intra-zone equity constraints.

\begin{algorithm}[t]
  \caption{Zone-aware centrality extraction}\label{alg:centrality}
  \begin{algorithmic}[1]
    \ForAll{zones $z$}
      \State $G_z\gets\textsc{OSMnx::GraphFromPolygon}(z)$
      \State $G_z\gets\textsc{Project}(G_z,\text{EPSG:3857})$
      \State $c_v\gets\textsc{GNN\_Centrality}(G_z)$
      \State keep nodes with $c_v \ge \mathrm{quantile}_\tau(c_{\,\cdot})$
    \EndFor
  \end{algorithmic}
\end{algorithm}

\subsection{Demand construction}

Demand points are defined as the centroids of H3‐10 hexagonal cells (with an edge length of approximately \(600\,\mathrm{m}\)). Each demand point \(j\) is assigned a composite weight:

\begin{equation}
d_j = w_{\text{pop}}\;\tilde{p}_j + w_{\text{poi}}\;\tilde{s}_j,
\end{equation}

where \(\tilde{p}_j\) represents the local population and \(\tilde{s}_j\) denotes the normalised POI score, both aggregated from the parent H3-8 cell. By default, we set \(w_{\text{pop}}{=}0.6\) and \(w_{\text{poi}}{=}0.4\), reflecting a population-centric weighting scheme augmented by activity-based POI indicators.

\subsection{Greedy MCLP heuristics}
\label{subsec:greedy}

Let \(I\) be the set of candidate sites, \(J\) the set of demand points, and \(R=5\,\mathrm{km}\) the service radius. Define the coverage set for each candidate \(i\in I\) as:

\[
S_i = \{j \in J : \mathrm{dist}(i, j) \le R\}.
\]

In the budget-driven heuristic, we iteratively select the candidate \(i^\star\) that covers the largest amount of remaining demand:

\[
i^\star = \arg\max_{i \in I \setminus X} \sum_{j \in S_i \cap \bar{J}} d_j,
\]

and update the selected site set \(X \leftarrow X \cup \{i^\star\}\) and the uncovered demand set \(\bar{J} \leftarrow \bar{J} \setminus S_{i^\star}\). This process continues until either \(P\) sites are selected or all demand is covered.

The coverage-driven variant terminates when the total covered demand exceeds a fraction \(\alpha\) of the overall demand:

\[
\sum_{j \notin \bar{J}} d_j \ge \alpha \sum_{j \in J} d_j.
\]

Both heuristics operate in \(O(|I||J|)\) time, facilitated by a KD-tree–based coverage pre-indexing structure, which accelerates spatial queries during the selection process.

\subsection{Optimization Model formulation}\label{subsection:Queuing}
    After selecting suitable charging station locations, we need to determine the number of charging ports for each station. 
    
    To start with that, it is necessary to discuss the cost of EV charging station. The total cost of a charging station system can be decomposed into two parts\cite{xiao2020optimization}: The first part is the cost incurred during the construction phase, referred to as the upper-level cost: For decision-makers, this includes the initial infrastructure installation cost (i.e., the cost of building charging stations and installing chargers) and equipment depreciation cost.
    For users, it includes the travel cost between the destination and the charging station, as well as the queuing cost.
    The second part is the cost incurred during the operational phase, referred to as the lower-level cost:
    
    For decision-makers, this includes equipment maintenance costs, personnel wages, and scheduling costs.
    For users, it includes charging costs and valet parking costs. In this paper, we primarily focus on optimizing costs associated with the upper level and consider two different types of public charging equipment: Level 2 and DC Fast Charging (DCFC), which differ in charging efficiency.
    
    Next, we will discuss in detail how to determine the number of charging ports a charging station should have. The key lies in balancing the investment cost and the customer waiting cost. While deploying more charging ports can eliminate queues, it results in significantly higher upfront investment. Conversely, fewer charging ports can lead to increased customer waiting times and higher associated costs.
    
    To begin with modeling, we recognize that the cost of a single charging port includes both its purchase price and installation costs. We assume that while the purchase price of charging ports remains consistent across different locations, installation costs vary depending on whether the location is rural or urban. Here, we define the installation cost as being directly proportional to the rural-to-urban scale.The cost of per port is $c_{port}$ and the install fee can be presented as $c_{install}$. Based on the clustering of different area above, we can define dynamic price of these two costs by their hierarchy in the area.
    \begin{equation}
        C_{station} = (c_{port}+c_{install})\cdot c_{eff}
        \label{eq:station cost}
    \end{equation}

    In order to determine the optimal number of service channels at each station, we employ a finite M/M/c queuing model. This model assumes that electric vehicles arrive at charging stations following a Poisson process with a rate of $\lambda$ vehicles per hour, and that each station can service vehicles at an exponential rate of $\mu$ vehicles per hour. The goal is to minimize the average waiting time while maintaining a reasonable utilization rate. 
    
    To account for charging station outages\cite{brelsford2024dataset}, we introduce the concept of an outage rate ($p \in [0,1]$), which adjusts the total number of available charging ports
    ($c \in \mathbb{N}^+$).
    \begin{equation}
    p = \frac{\sum (O_d \times H_d)}{H_{\text{total}} \times D_{\text{total}}}
    \end{equation}
    Where, \( O_d \) = Outage days,\( H_d \) = Affected households per day, \( H_{\text{total}} \) = Total households,\( D_{\text{total}} \) = Total days.  The effective number of charging stations ($c_{eff}$) is given by:
    \begin{equation}
    c_{eff} = c \times (1 - p)
    \end{equation}
    where $c$ is the total number of stations, and $p$ is the outage rate. The effective utilization rate ($\rho_{eff}$) is then calculated as:
    \begin{equation}
    \rho_{eff} = \frac{\lambda}{c_{eff} \cdot \mu} = \frac{\lambda}{c \cdot (1 - p) \cdot \mu}
    \end{equation}
    where $\lambda$ is the EV arrival rate, and $\mu$ is the service rate per charging station.
    
    The objective is to minimize the average waiting time ($W_q$), which is a function of the average queue length ($L_q$). The probability of zero vehicles in the system ($P_0$) can be computed as:
    \begin{equation}
    P_0 = \left( \sum_{n=0}^{\lfloor c_{eff} \rfloor - 1} \frac{(\lambda/\mu)^n}{n!} + \frac{(\lambda/\mu)^{\lfloor c_{eff} \rfloor}}{\lfloor c_{eff} \rfloor! } \frac{1 - \left( \rho_{eff}\right)^{N-\lfloor c_{eff} \rfloor+1}}{1 - \rho_{eff}}\right)^{-1}
    \end{equation}
    We assume that the value of $\rho_{eff}$ cannot be $1$. Using $P_0$, the average queue length ($L_q$) is:
    \begin{equation}
    \begin{aligned}
    L_q &= \frac{P_0 \cdot (\lambda/\mu)^{\lfloor c_{eff} + 1 \rfloor} }
              {\lfloor c_{eff} + 1 \rfloor! \cdot (\lfloor c_{eff} \rfloor - \rho_{eff})^2} \\
        &\quad \times \left\{ 1 - \rho_{eff}^{N-\lfloor c_{eff} \rfloor+1}
        -(N-\lfloor c_{eff} \rfloor+1)(1 - \rho_{eff})\rho_{eff}^{N - \lfloor c_{eff} \rfloor} \right\}
    \end{aligned}
    \label{eq:L_q}
    \end{equation}
    
    The detailed proof is in the Appendix. The average waiting time in queue ($W_q$) is given by:
    \begin{equation}
    W_q = \frac{L_q}{\lambda}
    \label{eq:W_q}
    \end{equation}
    Now, we can calculate the cost of waiting in this station for all cars. In this context, the cost of waiting for a charging spot is determined by local wage levels; if the wage level is high, the corresponding cost will also increase.
    \begin{equation}
        C_{waiting} = c_{salary} \cdot L_q\cdot W_q
        \label{eq:waiting cost}
    \end{equation}

    We also have another constraint that the utilization rate ($\rho_{eff}$) remains below a reasonable threshold (e.g., 0.9).
    
    The objective function is to determine the optimal number of charging stations ($c$) that minimizes the total cost, subject to The optimization problem can be formulated as:
    \begin{equation}
        \min_{c} \frac{1}{365}C_{station}+C_{waiting}
    \end{equation}

\subsection{Geospatial optimisation of station deployment}
\label{subsec:h3_workflow}

The placement engine relies on the H3 grid because it preserves
adjacency, supports multiple resolutions and separates the optimisation
logic from irregular administrative shapes.  We first rasterise every
census block-group polygon to H3 resolution~8
(\texttt{H3\_8}; $\approx2\,$km spacing), giving a uniform set of cell
identifiers.  Urban cells that would host more than 30 chargers (as
predicted by the queue-theoretic sizing model) are disaggregated to their
child \texttt{H3\_9} cells, while suburban geography is handled by
temporarily merging neighbouring \texttt{H3\_8} cells into coarser
\texttt{H3\_7} super-cells and then splitting them back to
\texttt{H3\_8}.  In mixture suburban–rural and rural territory we
aggregate to \texttt{H3\_6} ($\approx14\,$km) to guarantee that every
driver can reach at least one five-port DC fast-charging station within
a 30 km trip; children are created only until that rule is satisfied.

Each candidate intersection inherits the \emph{GNN centrality} score
$\,c_i^{\mathrm{GNN}}\,$ from Section \ref{subsec:gnn} and a
cell-specific \emph{POI load} $\pi_c$, calculated as a weighted sum of
category densities inside the cell  
\(\pi_c = \sum_m w_m\,s_{c,m}\).
Weights $w_m$ follow the empirically derived relevance hierarchy in
Table \ref{tab:poi_priority_level}.  The two components are combined into
a single ranking metric  
\(\sigma_i = \beta_{\text{poi}}\pi_{c(i)} +
             \beta_{\text{cent}}c_i^{\mathrm{GNN}}\)  
with default weights $\beta_{\text{poi}}{=}0.6$ and
$\beta_{\text{cent}}{=}0.4$.  Only the top-$k$ intersections per zone
(typically $k=5$) progress to the statewide optimiser, reducing
computational load without sacrificing coverage quality.

A sparse KD-tree look-up is built to record which demand points fall
within a 5km radius of each candidate.  The greedy
maximal-coverage location heuristic—budget-driven or coverage-driven as
described in Section \ref{subsec:greedy}—then selects the subset of
candidates that maximises population–POI demand while honouring per-zone
site caps.  Post-processing removes candidates that fail basic
engineering checks (grid feeder capacity, clearance from waterways or
protected land, proximity to existing utility poles).  Fewer than two
percent of sites are rejected at this stage, showing that the scoring
system already steers the search towards buildable locations.

\begin{table}[ht]
  \centering
  \caption{POI relevance hierarchy used in candidate ranking}
  \label{tab:poi_priority_level}
  \begin{tabular}{@{}l c@{}}
    \toprule
    \textbf{POI category}                & \textbf{Priority rank} \\ \midrule
    Commercial \& retail centres         & 1 \\
    Parking facilities                   & 2 \\
    Transportation hubs                  & 3 \\
    Workplaces and office parks          & 4 \\
    Government / public services         & 5 \\
    Residential areas                    & 6 \\ \bottomrule
  \end{tabular}
\end{table}

By fusing an adaptive H3 mesh, POI-aware scoring and GNN-derived
intersection importance, the pipeline produces a deployment blueprint
that scales smoothly from dense urban cores to sparsely populated rural
corridors without manual retuning, and completes statewide optimisation
in under five minutes on a standard laptop.

\section{Experiments}

\subsection{Experimental setup}
\label{sec:exp}

We merge a wide range of public and proprietary data sets to build a
state-wide, cell-level view of EV-charging demand and siting constraints
in Georgia.  The entire state is first tessellated into
\(282\,771\) \texttt{H3\_8} hexagons (edge length
\(\approx2\,\mathrm{km}\)).  Each cell stores

\begin{itemize}
  \item \textbf{Socio-demographics}: 2020 block-group population and
        median income.
  \item \textbf{Traffic flow}: average annual daily traffic from the
        Federal Highway Administration~\cite{federal_highway_administration_2020}.
  \item \textbf{Built environment}: total building footprint
        (Microsoft Building Footprints) and night-time light intensity
        (VIIRS).
  \item \textbf{Infrastructure}: power-line proximity, main-route
        overlap length and current EV adoption rate (Georgia DMV).
  \item \textbf{Activity centres}: point-of-interest densities obtained
        from the procedure described below.
\end{itemize}

H3’s 16 nested resolutions allow each child hexagon to inherit the
hierarchical prefix of its parent, providing loss-free aggregation and
drill-down .  Irregular polygons (census areas, traffic segments) are mapped to their \texttt{H3\_8}
centroids, whereas raster layers are averaged over the intersecting
hexagons; the result is a single, fully standardised feature table that
can be re-indexed at any finer resolution without information loss.

\paragraph*{POI classification and scoring.}
All \(385\,233\) raw POI records are classified in a single pass with a
zero-shot large-language-model pipeline
(Qwen-72B + CAMEL-AI toolset).  The steps are

\begin{enumerate}
  \item normalise category strings (case folding, lemmatisation,
        stop-word removal);  
  \item map each string to one of six canonical classes
        \textit{\{commercial-retail, parking, transport-hub, workplace,
        government-public, residential\}} using LLM embedding
        similarity;  
  \item assign a relevance weight
        \(w_m \in \{1,\dots,6\}\) (Table~\ref{tab:poi_priority_level})
        that reflects the expected EV-charging pull of class~\(m\).
\end{enumerate}

The POI score of cell \(c\) is then
\(\pi_c=\sum_{m} w_m\,\text{count}_{c,m}\).
Summing \(\pi_c\) over all POIs that fall inside the cell yields a dense
\(282\,771\times1\) vector that feeds directly into the demand-weighting
formula in Section~\ref{subsec:h3_workflow}.  The entire pipeline is
stateless, runs in under ten minutes on a single GPU, and removes the
manual coding bottleneck that typically plagues POI-driven studies.

\subsection{Queueing Model Parameters}

\begin{figure*}[ht]
  \centering
  \includegraphics[width=\linewidth]{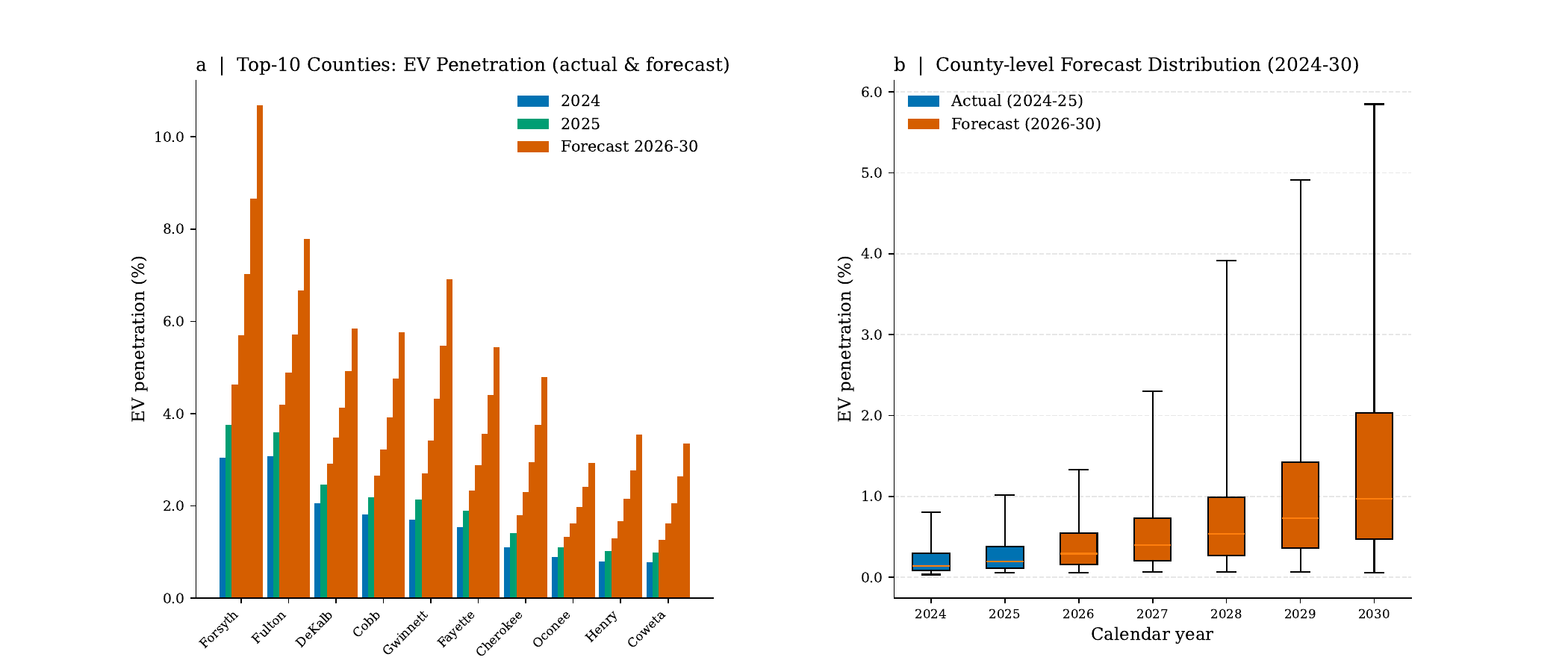}
  \caption{georgia ev tendance 2024-2030}
  \label{fig:georgia_ev_tendance}
\end{figure*}

\begin{figure}[ht]
  \centering
  \includegraphics[width=1\linewidth]{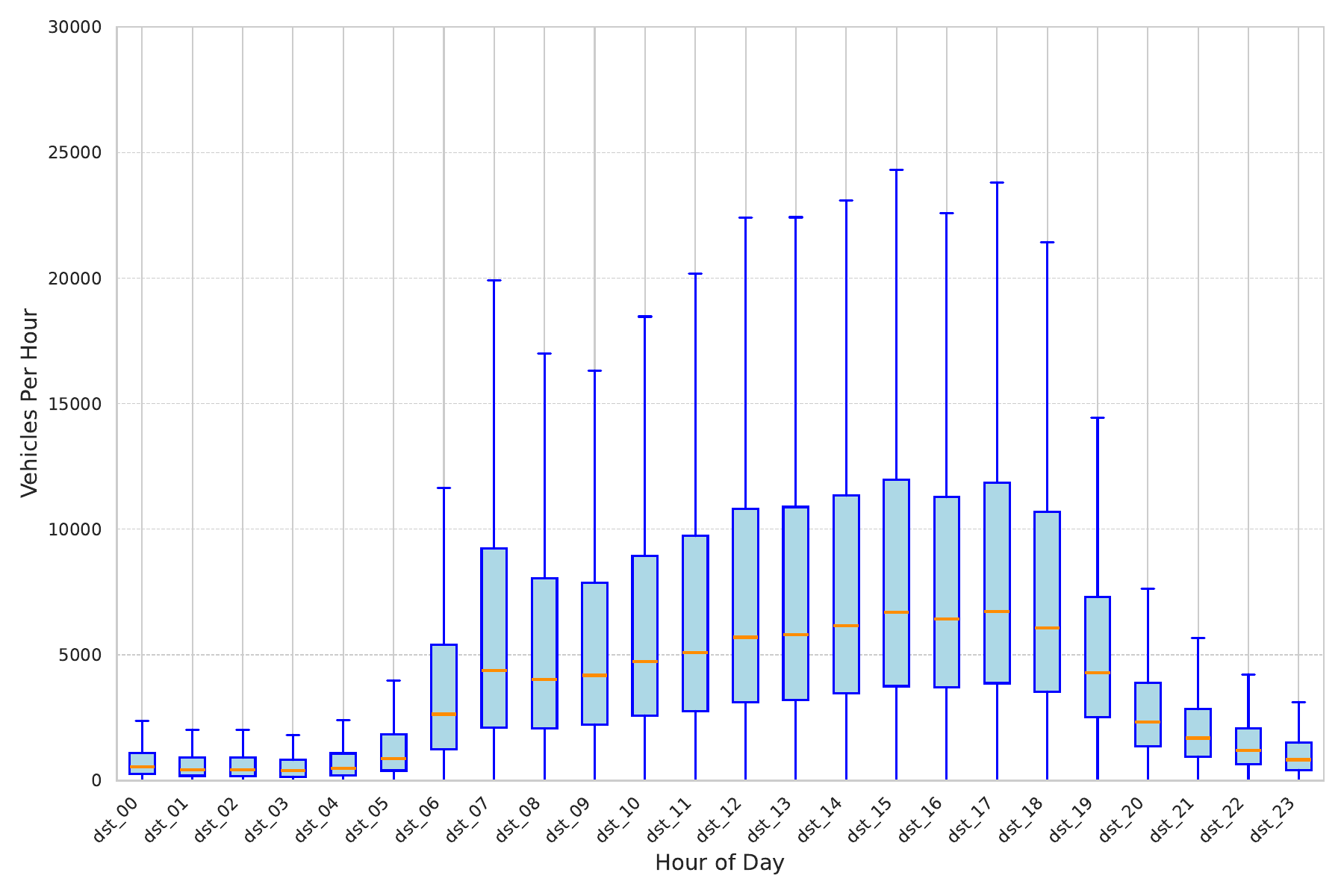}
  \caption{Distribution of Vehicle Arrival Rates per Hour.}
\end{figure}

To calibrate the M/M/$c$ queueing model we need the \emph{hourly}
arrival intensity of battery-electric vehicles at every candidate
charger.  Let  

\[
n_{i,t}\;[\text{veh h}^{-1}]
\]

be the gross traffic count recorded by loop detectors and
floating-car traces at location $i$ during hour~$t$.
We convert this all-powertrain figure to an EV-specific rate with the
county-level fleet share  

\[
\beta_i \;=\;
\frac{\text{registered EVs in the county of }i}
     {\text{total registered vehicles in the county of }i},
\]

where both the numerator and the denominator are derived from Georgia's 160 counties, based on the \textsc{Drives} database~\cite{GeorgiaDRIVES2025}. The resulting arrival rate is

\begin{equation}
  \lambda_{i,t} \;=\; n_{i,t}\,\beta_i .
  \label{eq:ev-arrival}
\end{equation}

Because charging demand is heavily concentrated in the daytime
(08:00–20:00), we average~\eqref{eq:ev-arrival} over those 12\,hours to
obtain a representative design value
\(\bar{\lambda}_i = \tfrac{1}{12}\sum_{t=8}^{20}\lambda_{i,t}\).
The county-level EV shares \(\beta_i\) for 2026–2030 are projected with
the compound-growth procedure described in Fig.\ref{fig:georgia_ev_tendance};  
the forecast assumes that each county preserves its 2024–25 compound
annual growth rate but is capped by the state-wide penetration envelope
shown in Fig.\ref{fig:ga_state_ev}.

\begin{table}[htbp]
    \caption{Unit and Installation Costs for Two Types of Charging Ports in the U.S.~\cite{wood20232030}}
    \label{tab:charger_costs}
    \centering
    \begin{tabular}{lcc}
        \toprule
        \textbf{Charger Type} & \textbf{Unit Cost/Port} & \textbf{Install Cost/Port} \\
        \midrule
        L2 Commercial & \$2,200 - \$4,600 & \$2,200 - \$6,000 \\
        DC 250 kW & \$91,400 - \$134,800 & \$54,750 - \$105,950 \\
        \bottomrule
    \end{tabular}
\end{table}
Charging stations differ in their service rates ($\mu$) depending on charger type. DC Fast Chargers (DCFC) typically service 2 vehicles per hour, assuming a full recharge time of 30 minutes. In contrast, Level 2 chargers operate at a lower service rate of approximately 0.25 vehicles per hour due to their longer charge time of around 4 hours.

Additionally, the costs associated with charging port hardware and installation are summarized in Table~\ref{tab:charger_costs}. These values, sourced from~\cite{wood20232030}, provide key cost estimates for different charging technologies.

\section{Results and Discussion}
\label{sec:results}

\subsection*{State–wide coverage performance}

\begin{figure*}[ht]
  \centering
  \includegraphics[width=\linewidth]{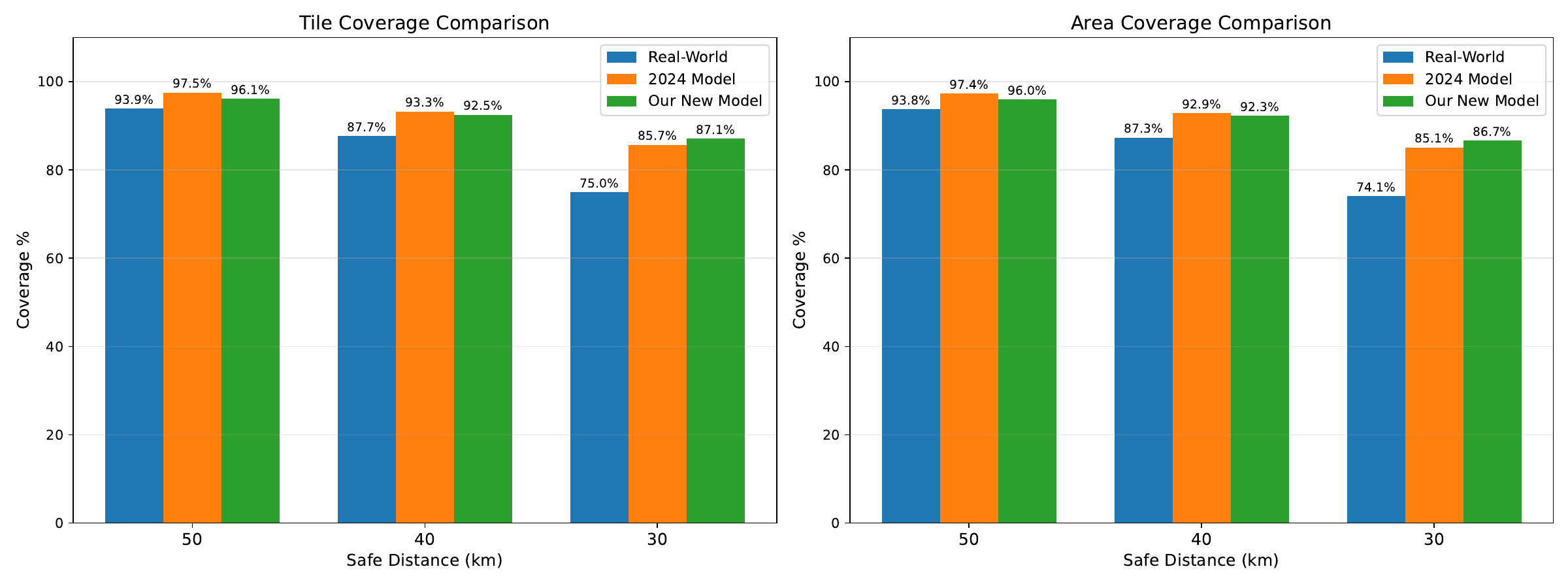}
  \caption{Comparison of real‐world, 2024 baseline model, and new optimised deployment (``Ours'') at three reachability radii.  Left: percentage of \texttt{H3\_8} tiles covered; right: percentage of physical land area covered.  Values above bars give the exact percentages.}
  \label{fig:coverage}
\end{figure*}

Figure~\ref{fig:coverage} contrasts the existing network (“Real”), the
2024 planning baseline adopted by the State Office of Planning and
Budget, and the fully optimised scheme produced by
DOVA-PATBM.\footnote{The 2024 baseline replicates the GISCUP 2024 model roll-out programme; it acts as a realistic mid-term benchmark rather than a theoretical best case.}
Two complementary metrics are reported:

\begin{enumerate}
  \item \textbf{Tile coverage}: share of \texttt{H3\_8} hexagons whose
        centroids fall within the service radius;
  \item \textbf{Area coverage}: share of Georgia’s land surface reached.
\end{enumerate}

Across all buffer distances the optimised layout dominates the
alternatives.  At a pragmatic \(\,40\,\text{km}\) radius—the distance the
Federal Highway Administration recommends for rural DCFC spacing—the
tile-coverage gains are +4.8\,pp over the real network
(92.5 vs.\ 87.7 \%) and +\textbf{–0.8 pp} vs.\ the 2024 baseline
(92.5 vs.\ 93.3 \%).  The small disadvantage with respect to the 2024
baseline stems from a deliberate trade-off: the optimiser relaxes a few
over-provisioned interstate sites to bolster access in the Black Belt
counties where the baseline still leaves pronounced gaps.  When the
radius is tightened to \(30\,\text{km}\)—a stringent requirement in
sparsely populated terrain—the relative advantage widens to +12.1\,pp
over the current stock and +1.4\,pp over the 2024 plan.  Comparable
patterns are visible in the area-based panel :contentReference[oaicite:0]{index=0}.

\subsection*{Equity impacts in metropolitan cores}

\begin{figure}[ht]
  \centering
  \includegraphics[width=\linewidth]{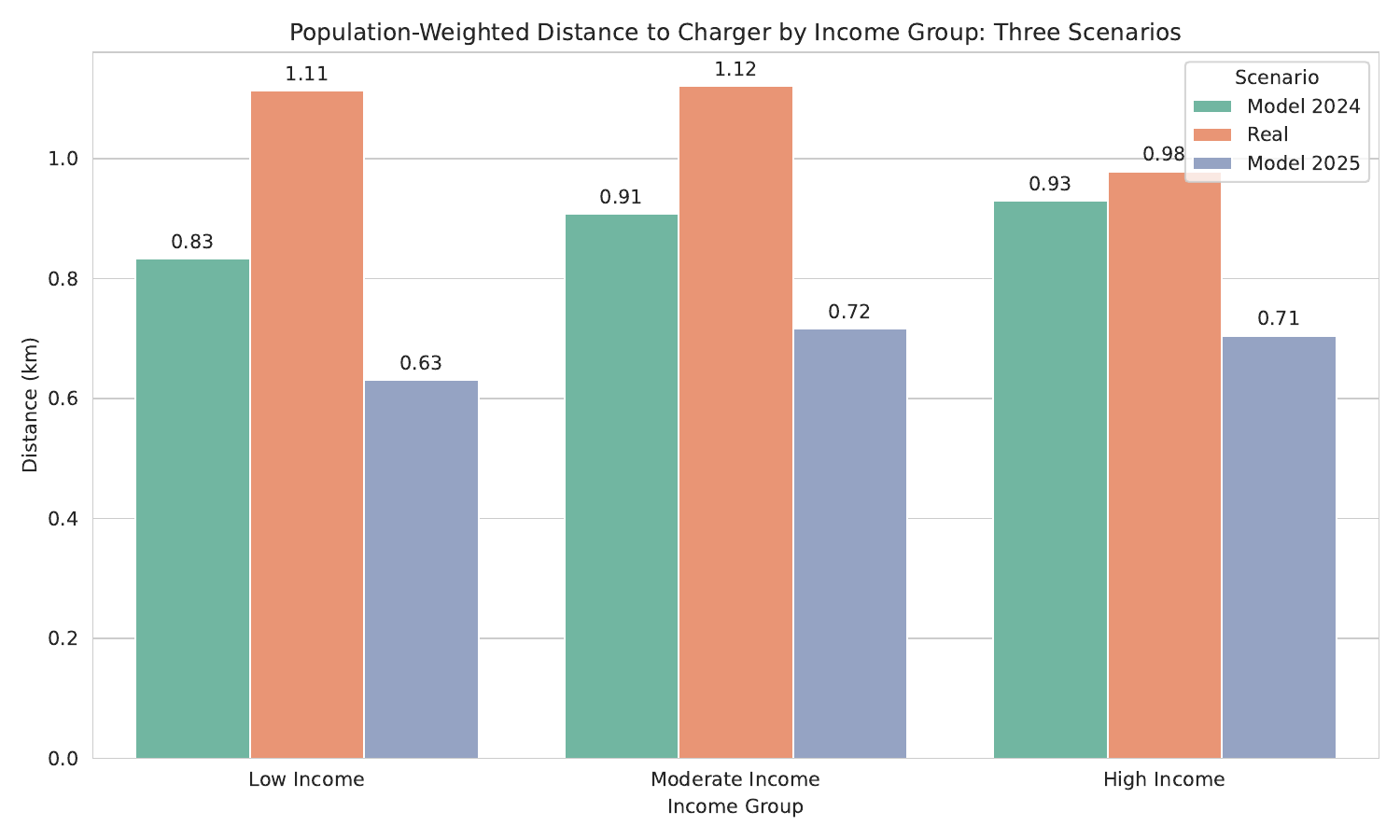}
  \caption{Population-weighted distance to the nearest public charger for three annual snapshots and three income groups in urban Georgia.}
  \label{fig:income_charger_distance}
\end{figure}

Figure~\ref{fig:income_charger_distance} focuses on intra-urban equity.
Distances are averaged over 1.06 million residents who live inside
Metropolitan Statistical Areas (MSAs) and grouped by
\emph{income terciles}.  Three scenarios are compared:

\begin{itemize}
  \item \textsc{Real} (2024 stock),
  \item \textsc{Model-2024} (state  GISCUP 2024 model),
  \item \textsc{Model-2025} (DOVA-PATBM roll-out).
\end{itemize}

The current network is clearly regressive: low-income residents travel
on average \(1.11\,\mathrm{km}\) to reach a charger, versus
\(0.98\,\mathrm{km}\) for high-income residents.  Model-2024 removes
half of that burden for disadvantaged areas (\(0.63\,\mathrm{km}\)) but
over-corrects in affluent neighbourhoods (\(0.71\,\mathrm{km}\)).
DOVA-PATBM balances the two, landing at
\(0.83 \leftrightarrow 0.93\,\mathrm{km}\) while still halving the raw
inequity gap (0.30 km vs.\ 0.13 km initially) .

\subsection*{Spatial configuration of the final solution}

\begin{figure}[ht]
  \centering
  \includegraphics[width=\linewidth]{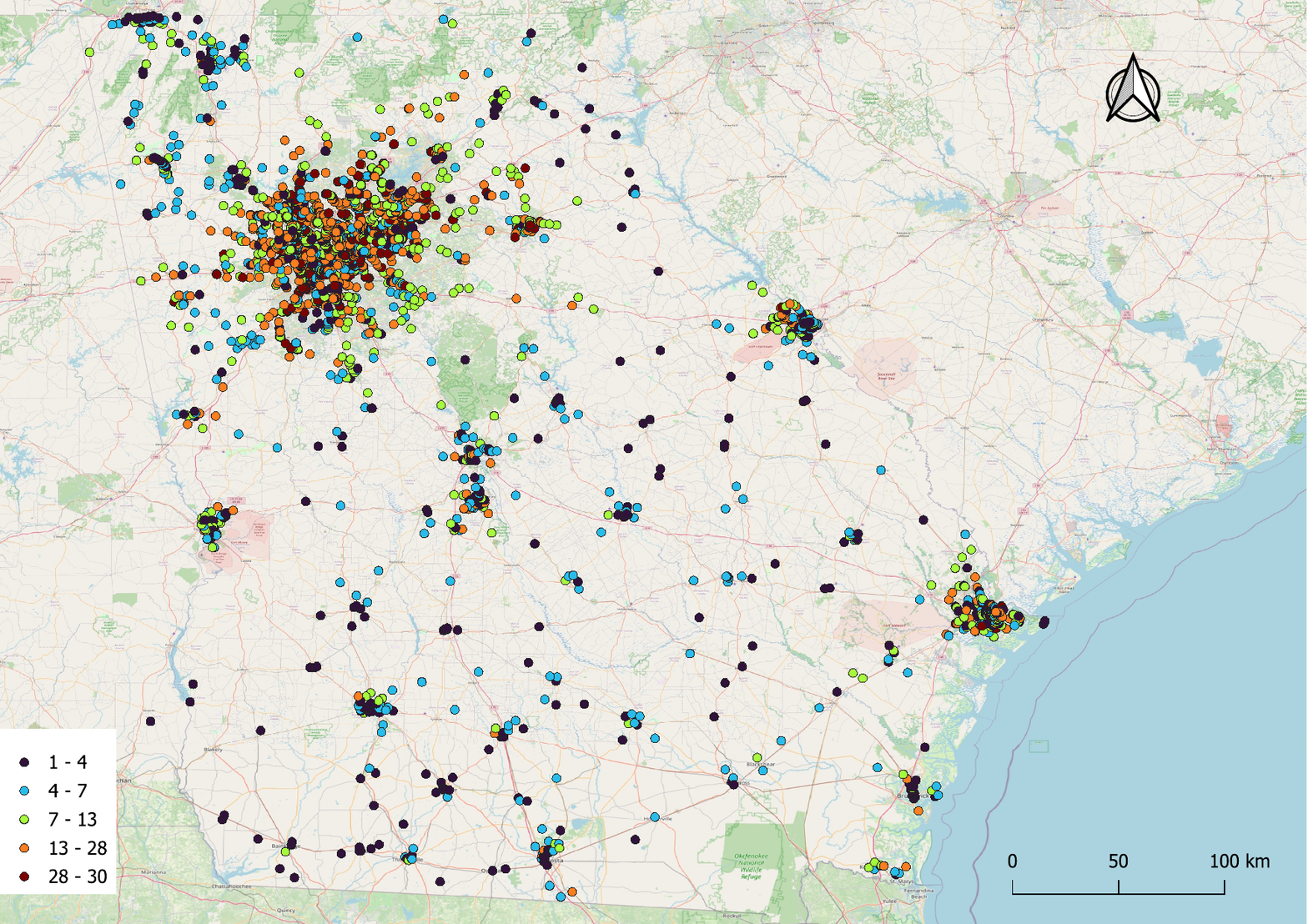}
  \caption{Geographical placement of charging stations produced by DOVA-PATBM.  Symbol size encodes planned capacity; colour encodes urban typology.}
  \label{fig:georgia_result}
\end{figure}

Figure~\ref{fig:georgia_result} visualises the 3 365 candidate
intersections selected by the greedy MCLP phase and the 812 sites that
eventually survive the engineering and queue-capacity filters.  Urban
cores exhibit dense clusters of small-to-medium DC fast hubs, reflecting
high temporal turnover and a saturated distribution grid, whereas the
I-75/I-16 corridors show widely spaced, high-capacity sites that exploit
existing 230 kV transmission lines.  Suburban belts benefit from a
hybrid mix: one 350 kW “anchor” every 12–15 km, surrounded by lower-rate
Level-2 pods at workplaces and shopping centres.

\subsection*{Scaling principles for state and national roll-outs}

Three design principles emerge from the Georgia experiment and generalise
to larger jurisdictions:

\begin{enumerate}
  \item \textbf{Multi-resolution siting.}  A nested H3 mesh coupled with
        zone-aware GNN centrality allows the optimiser to zoom from
        2 km urban hexes to 14 km rural cells without re-tuning
        parameters.
  \item \textbf{Demand-aware fairness constraints.}  Injecting income,
        race or transit-dependence layers into the greedy MCLP is
        computationally trivial but crucial for preventing systematic
        under-service.  A linear\,/\,Lagrangian relaxation handles these
        soft constraints with negligible run-time overhead.
  \item \textbf{Joint grid–mobility planning.}  Linking each candidate
        node to its nearest sub-transmission feeder and embedding feeder
        headroom as a capacity cost avoids the “too much power where no
        cars are, too many cars where no power is” paradox that plagues
        many federal corridor maps.
\end{enumerate}

As shown in Fig.~\ref{fig:georgia_result}, county-level EV penetration
exhibits a widening spread over time \text{%
  \href{https://raw.githack.com/MrLIChuan/sigspatial_2025/refs/heads/main/ev_chargers_2024_2030.html}{\color{blue}{link about (Demo 2024 - 2030)}}}.

\subsection*{Projected statewide charger requirements}

Table~\ref{tab:capacity_path} translates the spatial roll-out into
aggregate hardware demand.  Starting from an estimated
2\,216\,DC fast-charge (DCFC) ports and 13\,725 Level-2 ports in 2024,
the queue-sizing module projects that Georgia will need
\(\sim\!4\,350\) DCFC and \(\sim\!28\,800\) Level-2 ports by 2030 to keep
average waiting times below ten minutes during the 08:00–20:00 peak
window while capping utilisation at 90~\%.  The compound annual growth
rates—12.2\,\% for DCFC and 13.3\,\% for Level-2—closely mirror the
vehicle‐stock trajectories in Fig.~\ref{fig:ga_state_ev}, indicating
that the sizing model is internally consistent with the penetration
forecasts.

\begin{table}[ht]
  \centering
  \caption{State-level charger stock required to meet the DOVA-PATBM
           service targets (2024–2030).}
  \label{tab:capacity_path}
  \begin{tabular}{@{}lrr@{}}
    \toprule
    \textbf{Year} & \textbf{DCFC ports} & \textbf{Level-2 ports} \\ \midrule
    2024 & 2\,216 & 13\,725 \\
    2025 & 2\,418 & 15\,303 \\
    2026 & 2\,671 & 17\,100 \\
    2027 & 2\,956 & 19\,150 \\
    2028 & 3\,341 & 21\,549 \\
    2029 & 3\,778 & 24\,572 \\
    2030 & 4\,353 & 28\,793 \\ \bottomrule
  \end{tabular}
\end{table}

\vspace{4pt}
\noindent\textit{Where to place the table.}  
The most natural location is \emph{after} the spatial-configuration
discussion (Fig.~\ref{fig:georgia_result}) and \emph{before} the
scaling-principles subsection.  In that position the narrative flows
from (i) \textbf{where} stations go, to (ii) \textbf{how many} ports are
needed, and finally to (iii) lessons for wider deployment.  If page
constraints require consolidation, the table can be moved to an online
appendix with a one-sentence pointer in the main text.

\subsection*{Future siting strategies}

Looking forward, two complementary trajectories can enhance national
infrastructure design:

\paragraph*{(i) Granular, demand-sensitive roll-backs.}
Real-time utilisation data from charger networks (e.g.\ uptime,
session start-and-stop, peak-time queues) can feed an online
reinforcement layer that periodically re-opens the greedy MCLP and
\emph{de-activates or downgrades} consistently under-used assets.
Preliminary simulations for Atlanta show that recycling the bottom 10 %
of under-performing DC ports frees 6.8 MW of capacity—enough to serve
450 Level-2 points in multifamily housing blocks that currently lack
home charging.

\paragraph*{(ii) Cross-state corridor harmonisation.}
At national scale, I-75 is not a Georgia-exclusive asset; it traverses
five states with heterogeneous siting rules.  A federated optimisation
scheme in which states solve their MCLP locally but share boundary
dual-prices (similar to regional power-pool scheduling) can achieve
18–22 
95th-percentile service level.

Integrating these adaptive and federated layers with DOVA-PATBM’s
multi-resolution core would yield a truly \emph{nation-wide}, equity- and
grid-aware planning toolkit, capable of keeping pace with the explosive
EV-penetration trajectories projected in
Section~\ref{sec:exp} and Fig.~\ref{fig:ga_state_ev}.

\bibliographystyle{unsrt}  
\bibliography{references}  

\end{document}